%% file: ICRA2027_arxiv.tex
\documentclass[letterpaper, 10 pt, conference]{ieeeconf}  % Comment this line out if you need a4paper
\usepackage{graphicx}
\usepackage{color}
\usepackage{xcolor}

\usepackage[pagebackref,breaklinks,colorlinks,citecolor=green]{hyperref}
\usepackage{amsmath,amssymb}
\usepackage{multirow}
\usepackage{booktabs}
\usepackage{capt-of}
\usepackage[capitalize]{cleveref}

\crefname{section}{Sec.}{Secs.}
\crefname{table}{Tab.}{Tabs.}

\definecolor{warpcolor}{RGB}{157,57,246}
\definecolor{greencolor}{RGB}{10,197,10}

\IEEEoverridecommandlockouts  

\title{\LARGE \bf
\textcolor{warpcolor}{WARP}-VLA: \textcolor{warpcolor}{W}rist-Camera \textcolor{warpcolor}{A}daptation for View-\textcolor{warpcolor}{R}obust \textcolor{warpcolor}{P}olicy Execution in Vision-Language-Action Models
}
\author{%
Junmyeong Lee$^{1,\ast}$,
Dongmin Shin$^{1,2,\ast}$,
Min-Gyu Park$^{2}$,
Wooseok Jeon$^{1}$,
Inho Chang$^{2}$,
Hae-Gon Jeon$^{1,\dagger}$%
\thanks{$^{1}$Yonsei University.}%
\thanks{$^{2}$Korea Electronics Technology Institute(KETI).}%
\thanks{$^{\ast}$Equal Contribution.}%
\thanks{$^{\dagger}$Corresponding Author.}%
\\[0.5em]
{\href{https://june65.github.io/WARP-VLA}{%
\texttt{https://june65.github.io/WARP-VLA}}}%
}

\begin{document}
\IEEEaftertitletext{%
\begin{center}
    \centering
    \includegraphics[width=\textwidth]{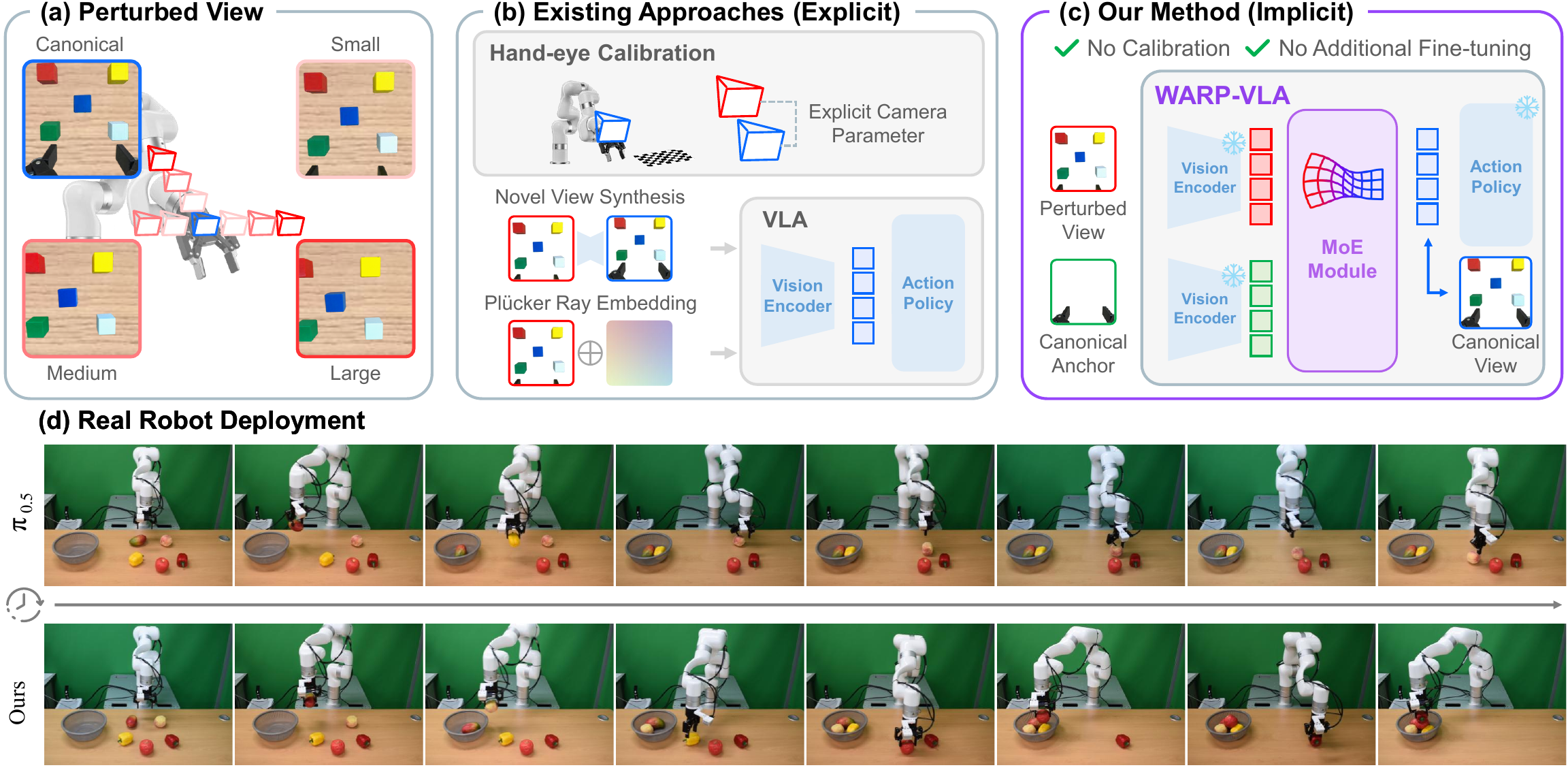}
    \captionof{figure}{\textbf{Overview of WARP-VLA.} (a) Changes in wrist-camera configuration alter interaction-relevant visual cues and degrade VLA performance. (b) Existing methods rely on novel view synthesis or explicit camera geometry to handle viewpoint changes. (c) Our method, WARP-VLA, aligns perturbed wrist camera features with the canonical feature space through an anchor-guided mixture of experts, enabling calibration-free and plug-and-play adaptation at deployment without further policy fine-tuning. (d) Real-robot results demonstrate the robustness of $\pi_{0.5}$ with WARP-VLA to wrist-camera perturbations.}
    \label{fig:teaser}
\end{center}%
}

\maketitle

\thispagestyle{empty}
\pagestyle{empty}
%%%%%%%%%%%%%%%%%%%%%%%%%%%%%%%%%%%%%%%%%%%%%%%%%%%%%%%%%%%%%%%%%%%%%%%%%%%%%%%%
\begin{abstract}
Despite recent advances in Vision-Language-Action models (VLAs) for robotic manipulation, their performance remains sensitive to changes in camera configuration. The problem becomes more evident in cross-setup deployment, as reproducing the exact camera pose used for training is nearly impossible. Unlike fixed external views, wrist views are more challenging because the camera moves with the robot, causing even small mounting variations to alter fine-grained geometric cues. To address this, we propose WARP-VLA, a camera-view-robust VLA for diverse wrist camera configurations. WARP-VLA adopts a Mixture-of-Experts (MoE) architecture where individual experts learn view-specific feature transformations, and a router combines them based on implicit view information. This allows the policy to be deployed without requiring camera extrinsic parameters as additional input. Through experiments on the LIBERO benchmark, WARP-VLA improves the average success rate of $\pi_{0.5}$ from 39.2\% to 78.3\% under wrist-view perturbations. The real-robot experiments further show that the feature-level adaptation learned in simulation successfully transfers to diverse deployment settings. 
To facilitate reproducibility and future research, we release our wrist viewpoint robustness benchmark and a plug-and-play implementation.
\end{abstract}

%%%%%%%%%%%%%%%%%%%%%%%%%%%%%%%%%%%%%%%%%%%%%%%%%%%%%%%%%%%%%%%%%%%%%%%%%%%%%%%%
\section{INTRODUCTION}
Recent advancements in Vision-Language-Action models (VLAs) have shown promising performance across various manipulation tasks~\cite{rt22023arxiv,kim2024openvla,pmlr-v305-black25a,bjorck2025gr00t,kim2025fine,shukor2025smolvla,pmlr-v305-wen25b,team2025gemini,li2024cogact}.
However, VLAs often fail to maintain their performance when deployed on real robots due to discrepancies between training and deployment environments.
Variations in camera configuration are particularly critical since visual observations directly determine the actions generated by VLAs.
In practice, reproducing the exact camera configuration during deployment is difficult because many robot datasets do not provide camera calibration information, including camera extrinsics and focal lengths. Even when such information is available, physical factors such as mounting tolerances and robot vibrations can introduce additional viewpoint changes during execution.

This problem is especially pronounced for wrist cameras, which provide fine-grained visual information required for precise manipulation~\cite{jangir2022look,pmlr-v164-valassakis22a}. 
Unlike external cameras that observe the workspace from fixed viewpoints, wrist cameras capture close-range views of the interaction region. 
Consequently, even small variations in wrist-camera configurations can alter interaction-relevant visual information, such as the spatial relationship between the gripper and objects, as depicted in~\cref{fig:teaser}(a). 
% deploy할 때마다 setting이 달라지고 정형화된 방법이 없다
Furthermore, reproducing the original camera configuration is more difficult for wrist cameras because their mounting arrangements vary across robot and camera hardware. 
Together, these challenges highlight the need for robustness to wrist-camera variations to ensure reliable deployment of VLA policies.

Prior work has explored improving VLA robustness to camera variations, primarily focusing on external cameras. Extending these approaches to wrist cameras is a natural starting point.
As shown in~\cref{fig:teaser}(b), these methods typically address viewpoint changes by incorporating camera geometry or synthesizing observations from target viewpoints.
For example, camera-conditioned methods encode camera poses through Pl\"ucker ray embeddings~\cite{jiang2026knowyourcamera}, while view synthesis approaches generate observations from the viewpoints used during policy training~\cite{heo2026anycamvla}. However, such approaches require accurate camera calibration or additional camera-specific models, which limits their applicability when the original wrist-camera configuration is unavailable. For wrist cameras, inaccurate calibration can also compromise the geometry-based conditioning or view synthesis used to interpret close-range interactions.

To this end, we introduce \textbf{W}rist-camera \textbf{A}daptation for view-\textbf{R}obust \textbf{P}olicy (\textbf{WARP}-VLA), a view-invariant VLA framework that implicitly learns and adapts diverse wrist camera view features through a Mixture-of-Experts (MoE) architecture, as illustrated in~\cref{fig:teaser}(c). WARP-VLA uses multiple experts to represent each camera-dependent transformation in wrist camera features observed during training. In a new camera setting, the framework combines nearby view experts through feature-conditioned routing weights, enabling smooth adaptation across view variations. Since WARP-VLA performs adaptation at the feature level, it provides a plug-and-play module applicable to other VLA policies trained in different simulation environments. This allows VLA policies to handle diverse wrist views in real-world deployment without additional policy fine-tuning.

To evaluate robustness to wrist-camera perturbations, we adapt the LIBERO~\cite{liu2023libero} and LIBERO-Plus~\cite{fei25libero} benchmarks to introduce a comprehensive wrist-view evaluation protocol covering diverse camera configurations. Through extensive experiments in simulation and on real robots, we demonstrate that WARP-VLA achieves strong robustness across diverse wrist-camera configurations while preserving performance under the original camera setup. Specifically, WARP-VLA generalizes across representative VLA backbones such as $\pi_{0.5}$ without requiring policy-specific modifications. Notably, applying WARP-VLA to $\pi_{0.5}$ improves the average success rate under wrist-camera perturbations from 39.2\% to 78.3\%. Furthermore, as shown in the real-world deployment results in~\cref{fig:teaser}(d), the baseline struggles to locate the object for grasping when the camera position changes, whereas WARP-VLA successfully grasps the object.

Our contributions are threefold:
\begin{itemize}
    \item We introduce a systematic benchmark for evaluating VLA robustness to wrist-camera variations, including a view-augmented dataset constructed from the LIBERO benchmark with diverse wrist-camera configurations.

    \item We propose WARP-VLA, a feature adaptation framework based on an MoE architecture that learns view-specific transformations and combines them through feature-conditioned routing without camera calibration.

    \item We demonstrate robust performance across diverse wrist-camera configurations and compatibility with different VLA backbones through extensive experiments in simulation and on real robots.
\end{itemize}

\section{RELATED WORK}

\subsection{View-Invariant VLAs}

Recent VLAs train a single policy on large-scale datasets to improve generalization across tasks and robot embodiments~\cite{rt22023arxiv,kim2024openvla,bjorck2025gr00t,rt12022arxiv,octo_2023,black2024pi}. Nevertheless, such policies assume similar settings during training and deployment, leaving them sensitive to different camera configurations. The fundamental challenge is that camera viewpoint changes alter not only visual appearance but also the geometric correspondence between observations and actions. 

A straightforward approach to improving viewpoint robustness is to expose policies to diverse camera views during training. VISTA~\cite{tian2024vista} and RoVi-Aug~\cite{chen2024roviaug} expand viewpoint coverage through synthesized demonstrations, reducing the need to collect additional demonstrations from different camera positions. These methods reduce multi-view data collection but leave policies to infer geometry from augmented observations. To provide explicit geometric cues,  KYC~\cite{jiang2026knowyourcamera} conditions policies on camera parameters, and StereoVLA~\cite{deng2025stereovla} incorporates stereo cues. OC-VLA~\cite{zhang2026grounding} and CamVLA~\cite{li2026fixed} further simplify visual-action correspondence by predicting camera-frame actions and converting them into robot-frame commands. However, limited training-view coverage remains a challenge. To avoid policy retraining, AnyCamVLA~\cite{heo2026anycamvla} instead uses novel view synthesis (NVS) to transform test-time observations into the policy's training viewpoints. This allows a frozen policy to handle camera changes without additional demonstrations or policy fine-tuning, but the NVS model still requires fine-tuning on multi-view images for the simulation domain.

\subsection{Geometry-Aware VLAs} %Geometry Foundation Models 

Incorporating geometric priors has been demonstrated to be an effective approach to improving VLA robustness to viewpoint changes. Explicit-geometry methods, including SpatialVLA~\cite{qu2025spatialvla}, GeoVLA~\cite{sun2025geovla}, PointVLA~\cite{li2026pointvla}, and Any3D-VLA~\cite{fan2026any3d}, introduce depth or point-cloud representations into the policy. However, making these cues useful also requires effective integration with action prediction ~\cite{yang2026understanding}. GeoAware-VLA~\cite{abouzeid2025geoaware}, G$^3$VLA~\cite{peng2026g}, and GAM~\cite{han2026geometric} incorporate geometric priors into visual encoding, cross-view fusion, or action decoding. To retain geometric knowledge without additional inference modules, Spatial Forcing~\cite{li2026spatial} instead aligns VLA features with geometric representations during training. These approaches make spatial information more accessible to the policy, while adapting perturbed wrist features to the canonical representations of a frozen VLA remains a distinct objective.

\section{THE PROPOSED METHOD}
% This section presents the MoE architecture and training procedure for wrist view invariant representation learning. We first introduce the overall MoE architecture for learning across different views in~\cref{sec:viewagnosic}. ~\cref{sec:problem} then formulate the adaptation objective as recovering canonical-view features using a pseudo-canonical anchor. Next, we describe how the router assigns each input view to the corresponding experts (\cref{sec:router}) and how they learn the view-specific transformations (\cref{sec:experts}). Finally,~\cref{sec:training} describes the training objective, improving the success rate of the downstream policy.

\subsection{Problem Definition}
\label{sec:problem}
We aim to adapt a VLA policy fine-tuned under a canonical wrist-camera configuration to perturbed wrist-camera configurations without explicit camera parameters during inference. Let $\mathbf{X}\in\mathbb{R}^{S\times D}$ denote the wrist camera tokens produced by an image encoder, where $S$ is the number of tokens and $D$ is the feature dimension. We denote the tokens from the canonical and perturbed views as $\mathbf{X}^{\mathrm{c}}$ and $\mathbf{X}^{\mathrm{p}}$. 

The adapter learns to transform perturbed-view features $\mathbf{X}^{\mathrm{p}}$ toward the canonical feature representation $\mathbf{X}^{\mathrm{c}}$:
\begin{equation}
    \hat{\mathbf{X}}^{\mathrm{c}}
    =
    f_{\phi}
    \left(
        \mathbf{X}^{\mathrm{p}},
        \mathbf{A}
    \right),
\label{eq:feature_recovery}
\end{equation}
where $f_{\phi}$ denotes the proposed view adapter parameterized by $\phi$. During training, $f_{\phi}$ is optimized to recover canonical-view features by minimizing the difference between $\hat{\mathbf{X}}^{\mathrm{c}}$ and the target canonical features $\mathbf{X}^{\mathrm{c}}$, where $\hat{\mathbf{X}}^{\mathrm{c}}$ is fed into the action policy network instead of $\mathbf{X}^{\mathrm{p}}$ to retain the robustness under camera configuration changes. 

To provide a reference representation of the canonical view, we introduce a pseudo-canonical feature anchor $\mathbf{A}\in\mathbb{R}^{S\times D}$. We compute the anchor $\mathbf{A}$ by averaging wrist-view features over all frames from the base policy's fine-tuning dataset, which contains only canonical-view observations. Averaging features across diverse scenes and tasks reduces observation-specific variations and provides a stable reference associated with the canonical camera configuration. 

\subsection{Canonical-View Adaptation with a Mixture of Experts}
\label{sec:viewagnosic}
\begin{figure}[t]
\centering
\includegraphics[width=\linewidth]{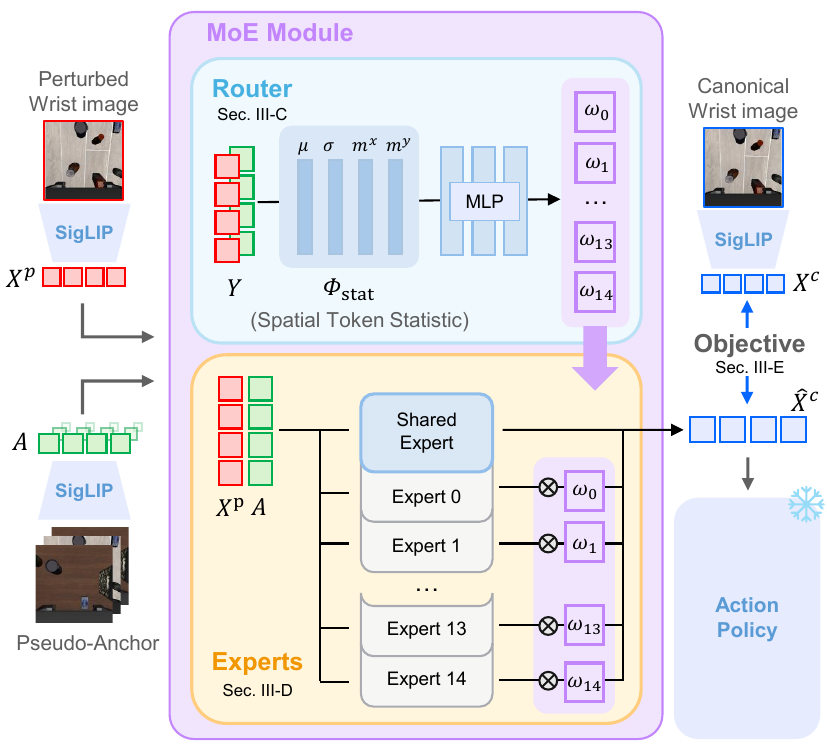}
\caption{\textbf{Training pipeline of WARP-VLA}.
The MoE adapter recovers canonical wrist features using a pseudo-anchor and expert weights predicted from spatial token statistics.
Paired canonical features supervise training, while the vision encoder and action policy remain frozen.}
\label{fig:structure}
\end{figure}

We adapt the base policy to wrist-camera perturbations by mapping perturbed-view features to the canonical-view feature space. As shown in~\cref{fig:structure}, an MoE architecture~\cite{jacobs1991adaptive} models mappings across the continuous wrist-camera configuration space using a finite set of learned transformation experts. A router combines their outputs based on input features without explicit camera parameters.

Let $F_e$ denote the transformation learned by expert $e$. The adapted wrist representation is obtained by combining the expert outputs:
\begin{equation}
F_{\mathrm{MoE}}(\mathbf{X}^{\mathrm{p}}, \mathbf{A})
=
\sum_{e=0}^{E-1}
w_eF_e(\mathbf{X}^{\mathrm{p}}, \mathbf{A}),
\label{eq:router_fusion}
\end{equation}
where $E$ denotes the number of experts.
The resulting representation replaces the original wrist camera tokens before entering the policy backbone.

We insert the adapter between the image encoder and the action policy of the policy backbone, performing adaptation in the image latent space. The adapter transforms wrist-camera features while preserving the token shape expected by the policy backbone and retaining the existing image encoder. This design allows the adaptation architecture to be applied to policies with compatible feature interfaces. We train a separate adapter for each VLA backbone and reuse the adapter trained on LIBERO in real-robot experiments without further adapter training.

\subsection{Anchor-Guided Routing with Spatial Token Statistics}
\label{sec:router}

The router determines how the view-specific experts are combined for a given wrist camera configuration. Since each expert learns a transformation for a specific region of the camera-configuration space, the router predicts the relative contribution of experts according to the deviation of the observed view from the canonical representation.

\begin{figure*}[ht]
\centering
\includegraphics[width=\linewidth]{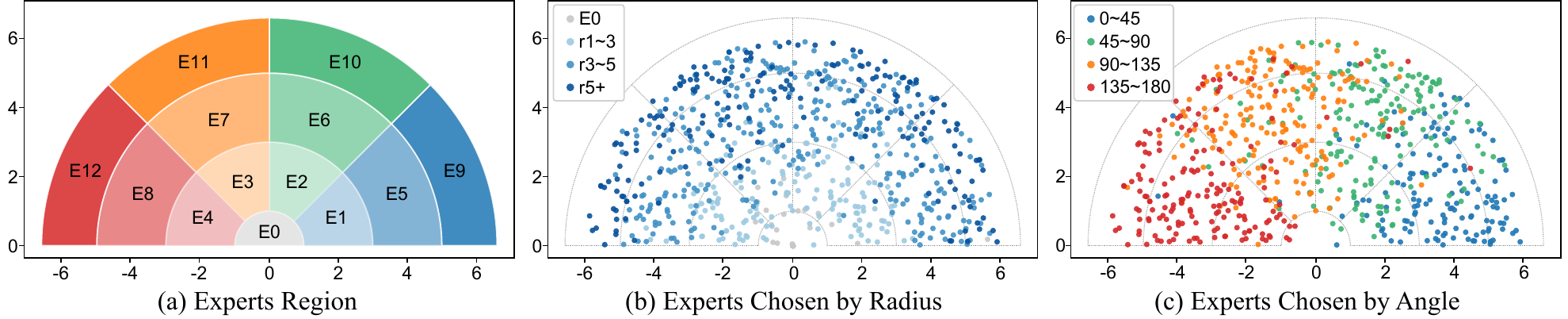}
\caption{Visualization of expert regions and routing assignments. (a) 13 target regions defined by radial distance and angle. (b) Router-selected radial groups. (c) Router-selected angular groups. Point locations represent ground-truth camera offsets, while colors indicate routing predictions.}
\vspace{-3mm}
\label{fig:router}
\end{figure*}

Given the perturbed-view features $\mathbf{X}^{\mathrm{p}}$ and the canonical anchor $\mathbf{A}$, we define the router input $\mathbf{z}$ as follows:

\begin{equation}
\mathbf{z}
=
\Phi_{\mathrm{stat}}(\mathbf{X}^{\mathrm{p}}-\mathbf{A})
\oplus
\Phi_{\mathrm{stat}}(\mathbf{A})
\in\mathbb{R}^{8D}.
\label{eq:router_representation}
\end{equation}
This residual statistical representation enables the router to capture how the observed wrist view deviates from the canonical configuration. During training, we randomly mix the anchor $\mathbf{A}$ with task-specific canonical anchors to prevent the router from overfitting to a specific canonical view.

For a feature tensor $\mathbf{Y}\in\{\mathbf{X}^{\mathrm{p}}-\mathbf{A},\,\mathbf{A}\}$, we extract channel-wise activation statistics and spatial token statistics:
\begin{equation}
\Phi_{\mathrm{stat}}(\mathbf{Y})
=
\boldsymbol{\mu}(\mathbf{Y})
\oplus
\boldsymbol{\sigma}(\mathbf{Y})
\oplus
\mathbf{m}^{x}(\mathbf{Y})
\oplus
\mathbf{m}^{y}(\mathbf{Y})
,
\label{eq:stat_representation}
\end{equation}
where $\boldsymbol{\mu}(\mathbf{Y})$ and $\boldsymbol{\sigma}(\mathbf{Y})$ denote the channel-wise mean and standard deviation, $\mathbf{m}^{x}(\mathbf{Y})$ and $\mathbf{m}^{y}(\mathbf{Y})$ denote the channel-wise first-order spatial moments, and $\oplus$ denotes concatenation. Instead of introducing an additional visual encoder, we derive the routing signal from the features already produced by the frozen policy. Regression analysis indicates that the spatial statistics of these features provide cues for distinguishing camera-view changes along different perturbation axes. Based on this observation, we compare the statistical representations of the perturbed wrist features and the pseudo-canonical anchor to construct the routing representation. This design allows the router to capture view-dependent feature shifts while reducing its reliance on scene- and task-specific visual content.

We apply LayerNorm to $\mathbf{z}$ before routing. The normalized representation is passed to the routing network:
\begin{equation}
\boldsymbol{\ell}
=
f_{\mathrm{gate}}
(
\operatorname{LN}(\mathbf{z})
)
\in\mathbb{R}^{E},
\label{eq:router_logits}
\end{equation}
where $f_{\mathrm{gate}}$ is a two-layer MLP with GELU activations followed by a linear projection, and $\ell_e$ denotes the routing logit of expert $e$. Consequently, the routing weight is computed as:

\begin{equation}
w_e
=
\frac{1}{G}
\frac{\exp(\ell_e)}
{\sum_{j\in\mathcal{E}_{g(e)}}\exp(\ell_j)},
\label{eq:group_routing}
\end{equation}
where $g(e)$ denotes the group containing expert $e$ and $G$ is the number of groups.

We use $E = 15$ experts, partitioned into $G = 2$ disjoint groups $\{\mathcal{E}_g\}_{g=1}^{G}$ according to the perturbation factor each group covers.
As shown in~\cref{fig:router}(a), the translation group contains 13 experts, one for each region of the semicircular $(r,\theta)$ offset space defined by radial distance and azimuth angle.
The tilt group contains the remaining 2 experts, which divide the tilt range.
The routing logits are normalized independently within each group, so the router predicts one distribution over the translation experts and another over the tilt experts.
This factorized routing allows the experts to represent combinations of translation and tilt variations while preventing competition among unrelated view factors.

\subsection{View-Specific Experts with Low-Rank Factorization}
\label{sec:experts}

Given perturbed wrist camera tokens $\mathbf{X}^{\mathrm{p}}$ and the pseudo-canonical anchor $\mathbf{A}$, each expert applies its factorized cross-attention blocks and produces a canonical-view feature:
\begin{equation}
F_e(\mathbf{X}^{\mathrm{p}},\mathbf{A})
=
\operatorname{Transformer}_e
(\mathbf{X}^{\mathrm{p}},\mathbf{A}),
\label{eq:expert_output}
\end{equation}
where the expert-specific behavior is determined by the low-rank residual parameters. The final adapted representation is obtained by combining the expert outputs using the routing weights described in~\cref{sec:router}.

Each expert learns a transformation from the observed wrist-view features to the canonical-view feature space. A naive design would assign an independent Transformer to each expert. However, wrist camera variations induce related feature shifts, where different views share a common transformation while requiring view-dependent corrections. We therefore factorize each expert into a shared transformation and a compact expert-specific residual. The shared component captures the common feature adaptation across views, while the residual component specializes the transformation for each view region.

We implement the shared transformation using a cross-attention Transformer architecture and introduce expert-specific low-rank residuals into its projection matrices inspired by LoRA~\cite{hu2022lora}. Each expert takes the perturbed wrist view tokens as queries and the pseudo-canonical anchor as keys and values, allowing the model to recover canonical-view features by attending to the reference representation. This design avoids replicating the full Transformer for every expert and enables scalable specialization through lightweight expert-specific residuals.

Let $W_{m}^{l}\in\mathbb{R}^{d_{\mathrm{in}}\times d_{\mathrm{out}}}$ denote the shared projection matrix of module $m$ in the $l$-th Transformer block, where $m$ corresponds to the attention and feed-forward projections. For expert $e$, the projection is defined as
\begin{equation}
W_{e,m}^{l}
=
W_{m}^{l}
+
A_{e,m}^{l}B_{e,m}^{l},
\label{eq:expert_factorization}
\end{equation}
where $A_{e,m}^{l}$ and $B_{e,m}^{l}$ are expert-specific low-rank factors with rank $r$. Since $r$ is smaller than the projection dimensions, each expert adds a small number of additional parameters while maintaining a dedicated view transformation.

To stabilize training, we initialize the expert-specific residual factors such that all experts initially share the same cross-attention transformation. Specifically, $A_{e,m}^{l}$ is randomly initialized while $B_{e,m}^{l}$ is initialized to zero. Therefore, all experts start from the same canonical-recovery function and gradually specialize to different view regions during training.

\subsection{Training Objective}
\label{sec:training}
The overall training objective combines an expert reconstruction loss and a routing supervision loss:
\begin{equation}
\mathcal{L}
=
\mathcal{L}_{\mathrm{e}}
+
\mathcal{L}_{\mathrm{g}},
\label{eq:training_loss}
\end{equation}
where $\mathcal{L}_{\mathrm{e}}$ denotes the expert reconstruction
loss and $\mathcal{L}_{\mathrm{g}}$ denotes the routing supervision
loss. These losses encourage expert specialization
and improve view-aware routing.

The experts are trained using a responsibility-weighted reconstruction loss, while the router is trained to match the estimated expert responsibilities:
\begin{equation}
\mathcal{L}_{\mathrm{e}}=\frac{1}{G}\mathbb{E}\left[\sum_e h_e d_e\right],
\,
\mathcal{L}_{\mathrm{g}}=-\frac{1}{G}\mathbb{E}\left[
\sum_e h_e\log g_e
\right].
\label{eq:expert_gate_loss}
\end{equation}

For each expert $e$, we compute its mean squared reconstruction error over all $S$ tokens and $D$ feature dimensions:
\begin{equation}
d_e
=
\frac{1}{SD}
\left\lVert
F_e(\mathbf{X}^{\mathrm{p}},\mathbf{A})
-
\mathbf{X}^{\mathrm{c}}
\right\rVert_F^2.
\label{eq:expert_error}
\end{equation}
Rather than computing a reconstruction error from the mixed expert output, we evaluate each expert independently to measure its ability to recover the canonical-view feature representation. This expert-level supervision encourages each expert to specialize in feature correction within its assigned region of the camera-configuration space, while preserving the spatial correspondence required for feature adaptation.

Inspired by teacher-directed learning~\cite{ebrahimpour2008view}, we estimate the responsibility of each expert using its routing probability, reconstruction quality, and training-time pose prior:
\begin{equation}
h_e =
\frac{g_e T_e \mathrm{e}^{-d_e/(2t^2)}}
{\sum_{j\in\mathcal{E}_k} g_j T_j \mathrm{e}^{-d_j/(2t^2)}},
\quad e\in\mathcal{E}_k,
\label{eq:expert_responsibility}
\end{equation}
where $\mathcal{E}_k$ denotes the set of experts in group $k$, $g_e=G w_e$ is the group-normalized routing probability, $T_e$ is the soft teacher mask derived from the known camera perturbation, and $t$ is a temperature parameter controlling sensitivity to reconstruction error. The responsibilities $h_e$ are treated as constants during gradient computation. The teacher mask uses overlapping pose bins, allowing neighboring experts to share responsibility near the boundaries between bins.

The teacher pose prior and canonical targets are used only as training-time supervision and are not required during inference. At inference time, the router predicts expert weights solely from the residual representation of the input wrist view features. This design enables plug-and-play adaptation of frozen VLA policies, requiring neither policy fine-tuning nor additional camera parameters.

\input{table/main_table.tex}

\section{Experimental results}

%To assess both viewpoint robustness and practical applicability, we evaluate WARP-VLA in simulation environments and on a real robot. We first describe the simulation and experimental setup and the construction of view-augmented training data in~\cref{sec:exp_details}. We then present simulation experiments evaluating viewpoint robustness, cross-domain generalization, and plug-and-play applicability across VLA backbones in~\cref{sec:sim_exp}. Next, we validate WARP-VLA under different wrist camera configurations on a real robot in~\cref{sec:real_exp}. Finally, we analyze the contributions of the routing and expert designs through ablation studies in~\cref{sec:ablation}.

\subsection{Implementation Details}
\label{sec:exp_details}

% We construct a wrist-camera robustness benchmark on LIBERO~\cite{liu2023libero} and LIBERO-Plus~\cite{fei25libero}, using a common protocol to evaluate VLA policies under controlled camera perturbations.~\cite{}

%\noindent\textbf{Training Data.}\quad 
We generate view-augmented training data using 2,000 LIBERO demonstrations covering 40 tasks across four suites. For each demonstration, we retain the canonical observations and re-render the trajectory under ten perturbed wrist-camera configurations. The mounting offsets are sampled uniformly at random within a semicircular translation region ($r\leq6\,\mathrm{cm}$, $\theta\in[0^\circ,180^\circ]$) and a tilt range ($\tau\in[0^\circ,10^\circ]$), relative to the canonical mount. Note that we only change the wrist-camera configuration, while all other settings remain fixed.

%\noindent\textbf{Implementation Details.}\quad 
We implement the shared expert using an eight-block ViT~\cite{dosovitskiy2020image} with a hidden dimension of 2,048 and a 256-dimensional attention module with four heads. We use 15 routed experts with rank-16 low-rank parameterization. The router maps a 16,384-dimensional pooled feature to routing weights through a two-layer MLP with a hidden dimension of 2048. We train WARP-VLA for 10,000 steps using Adam~\cite{kingma2014adam} with a batch size of 4,096 and a learning rate of $3\times10^{-4}$. We train our method on 8 NVIDIA RTX PRO 6000 GPUs for approximately 30 hours.

% 기존
% We generate training data from the LIBERO dataset~\cite{liu2023libero} by re-rendering demonstrations from multiple wrist camera configurations. Specifically, we use 2,000 demonstrations covering 40 tasks across the four suites. Each demonstration provides one canonical and ten re-rendered views with different wrist view camera mount offsets. The offsets are sampled uniformly at random for each demonstration over a semicircular translation region ($r\leq6\mathrm{cm}$, $\theta\in[0^\circ,180^\circ]$) and a tilt range ($\tau\in[0^\circ,10^\circ]$), relative to the canonical mount. Note that we only change the wrist view, while other configurations are fixed.

\subsection{Simulation Experiments}
\noindent\textbf{Simulation Benchmark.}\quad We evaluate all methods on LIBERO and LIBERO-Plus under four wrist-camera conditions: canonical, small, medium, and large. For each nonzero perturbation level $k\in\{\mathrm{small},\mathrm{medium},\mathrm{large}\}$, we define the camera-offset set as \begin{equation} \begin{aligned} \mathcal{O}_k &= \{r_k\} \times \{0^\circ,30^\circ,\ldots,180^\circ\} \times \{0^\circ,\tau_k\}, \\[2pt] (r_k,\tau_k) &= \begin{cases} (2\,\mathrm{cm},\,3^\circ), & k=\mathrm{small}, \\[1pt] (4\,\mathrm{cm},\,6^\circ), & k=\mathrm{medium}, \\[1pt] (6\,\mathrm{cm},\,9^\circ), & k=\mathrm{large}. \end{cases} \end{aligned} \label{eq:camera_offsets} \end{equation} Each nonzero level contains 14 camera configurations, combining seven azimuth angles with two tilt angles. For LIBERO, we evaluate one episode per configuration for each of the 40 tasks, yielding 560 episodes per perturbation level. Under the canonical condition, we use the unperturbed camera mount and the same 14 initializations per task used for the perturbed conditions. For LIBERO-Plus, we cycle through the 14 camera configurations in sequence over a total of 10K episodes per perturbation level. For each benchmark, all methods follow the same evaluation protocol. We report task success rates using the binary success criterion. All other evaluation settings follow the benchmark protocols.

\noindent\textbf{Baselines.}\quad Under this protocol, we compare WARP-VLA against backbone, geometry-aware, and camera-aware baselines. The backbone baselines are $\pi_{0.5}$~\cite{pmlr-v305-black25a} and GR00T N1.7~\cite{bjorck2025gr00t}. The geometry-aware baselines are Spatial Forcing~\cite{li2026spatial} and GAM~\cite{han2026geometric}, while the camera-aware baselines are KYC~\cite{jiang2026knowyourcamera} and AnyCamVLA~\cite{heo2026anycamvla}. To evaluate KYC under wrist-camera perturbations, we extend its camera-conditioning pathway to the wrist-camera stream of $\pi_{0.5}$. At each timestep, we update the camera extrinsics by composing the current end-effector pose with the fixed wrist-camera mount transform. For AnyCamVLA, we use the official $\pi_{0.5}$ checkpoint fine-tuned on LIBERO together with the released LVSM~\cite{jin2024lvsm} model adapted to the same simulation domain.

\begin{figure}[t]
\centering
\includegraphics[width=\linewidth]{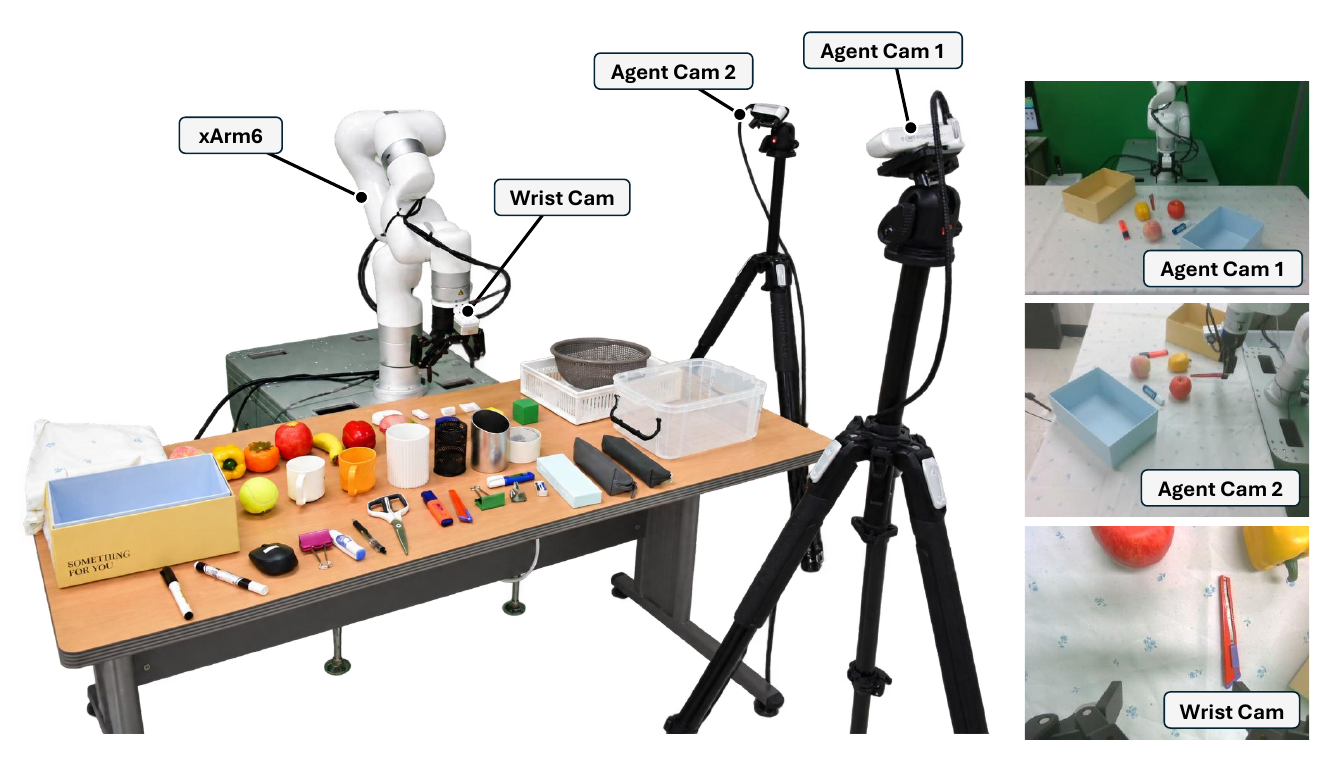}
\caption{Real-robot setup, showing the manipulation workspace and example observations from the external and wrist cameras.
}
\vspace{-3mm}
\label{fig:real_env}
\end{figure}

\label{sec:sim_exp}
\noindent\textbf{In-Domain Evaluation on LIBERO.}\quad
\cref{main_table} summarizes the success rate across the canonical wrist camera setting and three levels of viewpoint perturbation. Backbones retain high success rates in the canonical and small conditions, but their performance decreases sharply as the camera displacement increases. In particular, the average success rate of $\pi_{0.5}$ drops to 39.2\%, revealing the strong sensitivity to changes in the wrist camera configuration. Geometry-aware VLAs also suffer from substantial performance degradation under large viewpoint changes. For example, GAM achieves 98.4\% under the canonical configuration but decreases to 33.4\% under large perturbations. Since the models are primarily grounded in the canonical camera configuration, their performance drops substantially when exposed to unseen wrist viewpoints. These results highlight the need for a dedicated adapter that handles wrist camera variations for reliable deployment.

WARP-VLA outperforms camera-aware baselines such as KYC and AnyCamVLA under wrist-camera perturbations without requiring camera parameters at inference. Although AnyCamVLA benefits from novel view synthesis under small perturbations, our model achieves a higher success rate at the small level. Moreover, in terms of the average success rate across perturbed levels, WARP-VLA achieves higher performance than AnyCamVLA and KYC by 6.0 and 16.0 percentage points. This robustness stems from directly adapting wrist camera observations in feature space rather than explicitly synthesizing the canonical view. By continuously interpolating view-specific transformations, WARP-VLA generalizes better to unseen camera configurations where NVS or ray-based methods may degrade.

% Although AnyCamVLA benefits from Novel View Synthesis (NVS) under small perturbations and achieves the high success rate of 93.6\%, its performance decreases to 73.8\% and 49.5\% at the medium and large levels, respectively. This trend suggests that synthesizing observations from the policy's training viewpoint becomes increasingly challenging as the source and target camera configurations diverge. In contrast, WARP-VLA maintains better performance at the small level with a success rate of 95.5\% and outperforms AnyCamVLA by 6.7\% and 10.0\% at the medium and large levels, respectively. This robustness stems from directly adapting wrist camera observations in feature space rather than explicitly synthesizing the canonical view. By continuously interpolating view-specific transformations, WARP-VLA generalizes better to unseen camera configurations where NVS or ray-based methods may degrade.

\begin{figure*}[t]
\centering
\includegraphics[width=\linewidth]{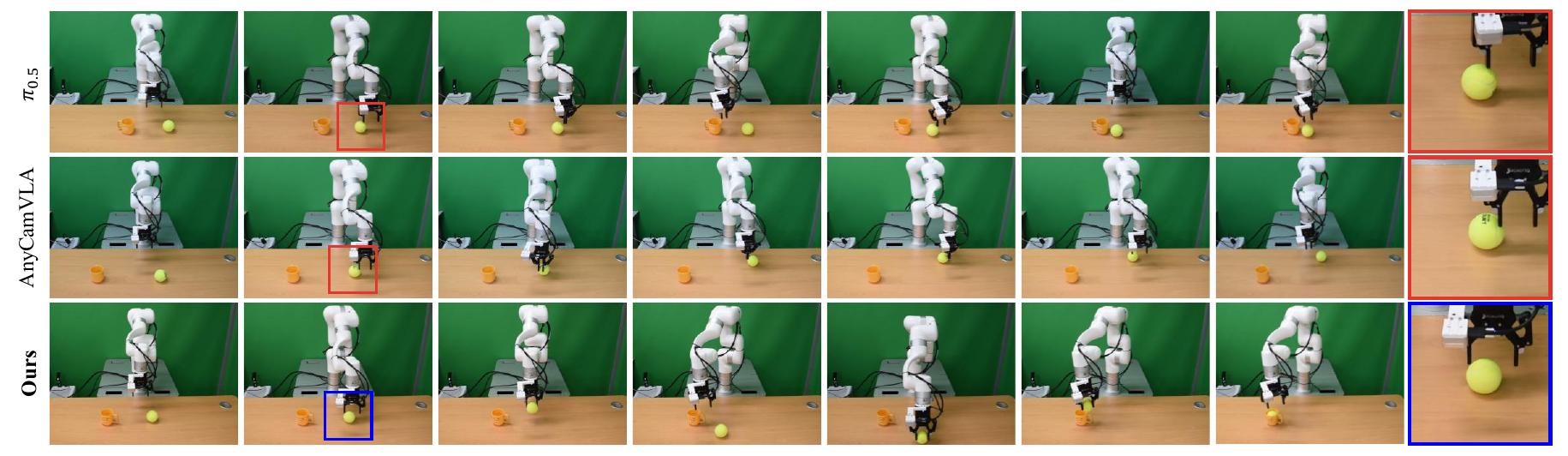}
\caption{Qualitative comparison of $\pi_{0.5}$, AnyCamVLA, and WARP-VLA under wrist-view perturbations on a real robot. The rightmost images show close-up views of the boxed regions, highlighting gripper–object misalignment in the baselines.}
\label{fig:qualitative}
\end{figure*}

\noindent\textbf{Out-of-Domain Generalization to LIBERO-Plus.}\quad
% pi에서 sota 
We evaluate the domain generalization of WARP-VLA by transferring the LIBERO-trained wrist view adapter to LIBERO-Plus without any additional fine-tuning. 
With $\pi_{0.5}$, WARP-VLA achieves the highest average success rate among the evaluated methods at 56.5\%, compared with 54.7\% for AnyCamVLA and 30.1\% for KYC. These results suggest that the camera-specific transformations learned on LIBERO remain effective across different scenes and wrist-camera configurations. By combining specialized experts in the pretrained visual feature space, WARP-VLA aligns perturbed wrist features with the representations expected by the frozen policy. The transferred adapters also improve the average success rates of $\pi_{0.5}$ and GR00T N1.7 by 27.7 and 19.1 percentage points, respectively. These gains support the effectiveness of the MoE adaptation architecture across different VLA backbones. This design enables the learned transformations to remain useful in the target domain without further adapter training or policy fine-tuning.

% AnyCamVLA performs well under small perturbations, where substantial overlap and limited visibility changes between the observed and canonical views may facilitate accurate view synthesis.

% \noindent\textbf{Plug-and-Play across VLA Backbones on RLBench.}\quad
% To evaluate the plug-and-play capability of WARP-VLA, we directly attach the view adapter trained on LIBERO to a frozen policy on RLBench without any additional training or domain-specific adaptation. In \cref{main_table2}, Applying the LIBERO-trained adapter to both $\pi_{0.5}$ and GR00T N1.7 improves their performance by ---\% and ---\%, respectively. By contrast, AnyCamVLA relies on a domain-specific LVSM and suffers from degraded performance when transferred to RLBench without retraining. This difference highlights the benefit of feature-space adaptation, as the pretrained latent space preserves a more consistent structure across domains and allows the LIBERO-learned transformation to generalize. 

\subsection{Real-Robot Experiments}
\label{sec:real_exp}

\noindent\textbf{Experimental Setup.}\quad 
In real-robot experiments, we compare WARP-VLA with $\pi_{0.5}$ and AnyCamVLA.
As shown in~\cref{fig:real_env}, we conduct real-robot experiments using a UFACTORY xArm6 robotic arm equipped with a Robotiq 2F-85 gripper. 
The robot is equipped with two RealSense D435 cameras as agent cameras and a RealSense D405 camera as the wrist camera. 
To evaluate robustness to wrist camera variations, we design and fabricate camera mounts with different wrist camera configurations using 3D printing. 
The mounts include one canonical configuration and nine perturbed configurations, with three separate variations at each perturbation level: x-axis translation, y-axis translation, and tilt. The small, medium, and large levels use translation magnitudes of 2, 4, and 6 cm along either axis and tilt angles of 3°, 6°, and 9°, respectively. 
Using the canonical camera mount, we collect 50 demonstrations for each of 10 tasks through teleoperation~\cite{welle2024quest2ros} with a Meta Quest Pro headset and use these demonstrations to fine-tune the baseline policies. We compute the pseudo-canonical anchor by averaging wrist-camera features over the canonical real-robot demonstrations, while keeping the LIBERO-trained adapter fixed. During evaluation, we replace the canonical mount with perturbed camera mounts and perform five trials for each task under each camera configuration.

\begin{figure}[t]
\centering
\includegraphics[width=\linewidth]{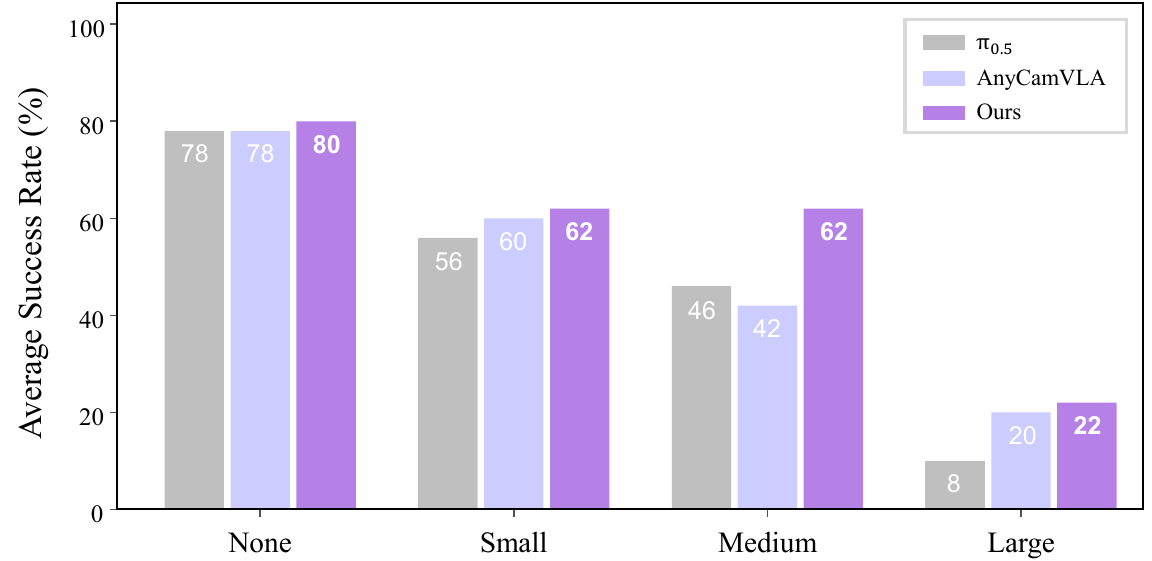}
\caption{Real-robot success rates (\%) across wrist-view perturbation levels.}
\label{fig:deploy}
\vspace{-5mm}
\end{figure}

\noindent\textbf{Quantitative Results.}\quad 
We evaluate the robustness of WARP-VLA under different wrist-camera configurations on the real robot.
For each camera configuration, we conduct five trials for each of the 10 tasks and report the average success rate. As shown in~\cref{fig:deploy}, both $\pi_{0.5}$ and AnyCamVLA achieve a 78\% success rate under the canonical camera configuration. Under wrist-camera perturbations, their average success rates decrease to 36.7\% and 40.7\%, respectively.
In contrast, WARP-VLA achieves higher success rates than both baselines across different wrist camera configurations, achieving 80\% under the canonical setting and 48.7\% under perturbed configurations. Compared with $\pi_{0.5}$ and AnyCamVLA, WARP-VLA improves the average success rate under camera perturbations by 12.0 and 8.0 percentage points, respectively. These results demonstrate that the proposed view adapter effectively reduces the sensitivity of VLA policies to wrist camera variations and improves robustness in real-world deployment.

\noindent\textbf{Qualitative Results.}\quad
\cref{fig:qualitative} compares grasping behavior under wrist-camera perturbations.
The $\pi_{0.5}$ and AnyCamVLA baselines approach the target but fail to center it between the gripper fingers, resulting in misplaced grasps.
These errors suggest insufficient compensation for wrist-view feature shifts.
The policies may follow the visual alignment learned under the canonical camera configuration, bringing the object in front of the wrist camera without correctly accounting for the changed camera--gripper relationship.
In contrast, WARP-VLA centers the object between the gripper fingers despite the perturbed camera configuration and completes the cup-and-ball task.
This comparison illustrates how feature-space view adaptation helps maintain the gripper and object alignment required for successful manipulation.

\subsection{Ablation Studies}
\label{sec:ablation}

\input{table/ablation}

\noindent\textbf{Multi-Expert Adaptation.}\quad
As shown in \cref{ablation}, we investigate whether a single shared adapter is sufficient for robust wrist view adaptation. We retain the shared ViT backbone while removing the router and expert-specific low-rank residuals, such that all camera configurations are handled by a single transformation. The resulting adapter improves upon the unadapted VLA, confirming that aligning perturbed wrist view features with the canonical feature space is itself beneficial. However, it remains less robust than WARP-VLA across perturbed views, indicating that a single shared transformation cannot adequately capture the heterogeneous feature changes induced by wrist camera displacement and tilt. These results support the use of view-specific experts for robust adaptation across diverse camera configurations.

\noindent\textbf{View-Conditioned Routing.}\quad
We isolate the contribution of input-dependent routing by retaining all view-specific experts while replacing the learned routing weights with uniform weights within each expert group. As shown in \cref{ablation}, uniformly combining the experts yields lower robustness than the learned routing scheme, even with the same expert capacity. This result suggests that the benefit of MoE comes not only from expert specialization but also from adapting their contributions to the observed wrist view. 
% The assigned experts achieve the minimum MSE across all ten factor bins, confirming consistent specialization along the radial, angular, and tilt dimensions.

\noindent\textbf{View-Space Structure of Learned Routing.}\quad
\cref{fig:router}(b, c) examines whether the learned router recovers the radial and angular structure used to define the experts. The left panel shows the 13 target regions defined over the $(r,\theta)$ camera-offset space, whereas the middle and right panels visualize actual samples colored according to the router-selected radial and angular groups. Although some overlap remains near the partition boundaries, the resulting assignments form clear concentric bands and angular sectors. Since the ground-truth offsets are not provided as router inputs, the alignment between the predicted assignments and target regions provides evidence that the wrist image tokens encode sufficient cues for view-dependent routing.

\section{Conclusion}
We have presented WARP-VLA, a mixture-of-view adapter for improving the robustness of vision-language-action policies against wrist camera variations. Unlike existing approaches relying on explicit camera representations, WARP-VLA learns to transform perturbed wrist camera observations into a canonical feature space using view-specific experts and an input-dependent router. We construct a systematic evaluation protocol with diverse wrist camera configurations and demonstrate that WARP-VLA substantially improves robustness across LIBERO and LIBERO-Plus benchmarks. Furthermore, real-robot experiments with physically perturbed camera mounts validate that the proposed approach generalizes to practical deployment scenarios. Our results highlight the importance of modeling wrist camera variations at the visual representation level for reliable deployment of VLA policies.

\noindent\textbf{Limitations.}\quad 
While WARP-VLA improves robustness to wrist-camera variations, its generalization remains constrained by the training camera distribution and the available visual evidence. Because adaptation operates on observed features, it cannot fully compensate when task-relevant regions are completely occluded or outside the field of view. One observed failure under large perturbations is premature gripper opening despite accurate object localization. Limited gripper visibility may contribute to this failure by making the grasp state difficult to infer. More reliable grasp-state estimation could therefore improve task success. Future work will expand camera coverage during training and explore temporal memory and complementary camera observations.

\addtolength{\textheight}{-12cm}   % This command serves to balance the column lengths

%%%%%%%%%%%%%%%%%%%%%%%%%%%%%%%%%%%%%%%%%%%%%%%%%%%%%%%%%%%%%%%%%%%%%%%%%%%%%%%%

%%%%%%%%%%%%%%%%%%%%%%%%%%%%%%%%%%%%%%%%%%%%%%%%%%%%%%%%%%%%%%%%%%%%%%%%%%%%%%%%



% \section*{ACKNOWLEDGMENT}

% The preferred spelling of the word ÒacknowledgmentÓ in America is without an ÒeÓ after the ÒgÓ. Avoid the stilted expression, ÒOne of us (R. B. G.) thanks . . .Ó  Instead, try ÒR. B. G. thanksÓ. Put sponsor acknowledgments in the unnumbered footnote on the first page.

% %%%%%%%%%%%%%%%%%%%%%%%%%%%%%%%%%%%%%%%%%%%%%%%%%%%%%%%%%%%%%%%%%%%%%%%%%%%%%%%%

% References are important to the reader; therefore, each citation must be complete and correct. If at all possible, references should be commonly available publications.

\bibliographystyle{IEEEtran}
\bibliography{IEEEfull}

\end{document}

%% file: table/main_table.tex
\begin{table*}[ht]
\vspace{3mm}
\caption{Success rates (\%) under wrist-view perturbations on LIBERO and LIBERO-Plus.
}
\setlength{\tabcolsep}{3pt} 
\label{main_table}
\begin{center}
% \begin{tabular}{c|cccc|c|cccc|c|cccc|c|cccc|c|c}
% \toprule
% \multirow{3}{*}{Baselines} & \multicolumn{21}{c}{LIBERO} \\
% \cmidrule{2-22}
%  & \multicolumn{5}{c|}{None} & \multicolumn{5}{c|}{Small} & \multicolumn{5}{c|}{Medium} & \multicolumn{5}{c|}{Large} & \multirow{2}{*}{Average} \\
% \cmidrule{2-21} 
%  & Spatial & Object & Goal & Long & Avg. & Spatial & Object & Goal & Long & Avg. & Spatial & Object & Goal & Long & Avg. & Spatial & Object & Goal & Long & Avg. & \\
% \midrule
% $\pi_0$ & 95.0 & 96.4 & 96.4 & 80.7 & 92.1 & 92.1 & 88.6 & 90.7 & 63.6 & 83.8 & 57.1 & 35.7 & 63.6 & 9.3 & 41.4 & 28.6 & 5.7 & 35.7 & 2.1 & 18.0 & 58.8 \\
% \bottomrule
% \end{tabular}
\scriptsize
\begin{tabular}{c|c|ccc|c|c|ccc|c}
\toprule
\multirow{3}{*}{Baselines\vspace{-10pt}}
& \multicolumn{5}{c|}{LIBERO (In-Domain)}
& \multicolumn{5}{c}{LIBERO-Plus (Plug-and-Play)} \\
\cmidrule{2-6} \cmidrule{7-11}
& \multirow{2}{*}{~Canonical~\vspace{-6pt}} & \multicolumn{4}{c|}{Perturbed} & \multirow{2}{*}{~Canonical~\vspace{-6pt}} & \multicolumn{4}{c}{Perturbed} \\
\cmidrule{3-6} \cmidrule{8-11}
&  & ~~~~Small~~~~ & ~~~Medium~~~ & ~~~~Large~~~~ & ~~~Average~~~ &  & ~~~~Small~~~~ & ~~~Medium~~~ & ~~~~Large~~~~ & ~~~Average~~~ \\    
\midrule
% $\pi_{0}$ & 92.1 & 83.8 & 41.4 & 18.0 & 47.7 & \textcolor{gray!60}{N/A} & \textcolor{gray!60}{N/A} & \textcolor{gray!60}{N/A} & \textcolor{gray!60}{N/A} & \textcolor{gray!60}{N/A} \\
% OpenVLA-OFT & 97.4 & 88.2 & 49.9 & 32.0 & 56.7 & \textcolor{gray!60}{N/A} & \textcolor{gray!60}{N/A} & \textcolor{gray!60}{N/A} & \textcolor{gray!60}{N/A} & \textcolor{gray!60}{N/A} \\
% \midrule
Spatial Forcing & 97.7 & 88.6 & 43.1 & 21.1 & 50.9 & \textcolor{gray!60}{N/A} & \textcolor{gray!60}{-} & \textcolor{gray!60}{-} & \textcolor{gray!60}{-} & \textcolor{gray!60}{-} \\
GAM & \textbf{98.4} & 87.5 & 55.0 & 33.4 & 58.6 & \textcolor{gray!60}{N/A} & \textcolor{gray!60}{-} & \textcolor{gray!60}{-} & \textcolor{gray!60}{-} & \textcolor{gray!60}{-} \\
\midrule
%AnyCamVLA' & 96.4 & 96.1 & 80.2 & 55.7 & 77.3   & - & 81.4 & 59.6 & 34.0 & 58.3* \\
%AnyCamVLA*& 93.8 & 93.8 & 75.0 & 52.3 & 73.7   & - &  &  &  &  \\
AnyCamVLA & 97.1 & 93.6 & 73.8 & 49.5 & 72.3 & 81.9 & \textbf{79.0} & 55.3 & 30.0 & 54.7 \\
%KYC &  &  &  &  &    &  &  &  &  &  \\
KYC& 92.3 & 86.8 & 60.0 & 40.2 & 62.3 & 75.5 & 64.7 & 18.6 & 6.9 & 30.1 \\
\midrule 
%$\pi_{0.5}'$ & 98.0 & 84.5 & 17.1 & 10.0 & 57.2 & - & 63.3 & 7.0 & 3.9 & 24.7* \\
GR00T N1.7 & 92.1 & 86.1 & 26.1 & 11.8 & 41.3 & 74.9 & 65.3 & 18.7 & 8.3 & 30.8 \\
%GR00T N1.7 + \textbf{Ours}
%\textbf{+Ours} (GR00T N1.7)
\textbf{+Ours} & 94.1 & 88.6\; \textcolor{greencolor}{+2.5} & 60.5\; \textcolor{greencolor}{+34.4} & 50.4\; \textcolor{greencolor}{+38.6} & 66.5\; \textcolor{greencolor}{+25.2} & 74.6 & 70.7\; \textcolor{greencolor}{+5.4} & 45.2\; \textcolor{greencolor}{+26.5} & 33.7\; \textcolor{greencolor}{+25.4} & 49.9\; \textcolor{greencolor}{+19.1} \\
\midrule
%$\pi_{0.5}$ + Ours' (no res) & 95.9 & 95.7 & 83.9 & 67.5 & 82.3 & - & 74.6 & 59.0 & 41.0 & 58.2* \\
$\pi_{0.5}$ & 95.2 & 90.5 & 20.9 & 6.1 & 39.2 & 82.7 & 72.2 & 11.2 & 2.9 & 28.8 \\
%\textbf{Ours} ($\pi_{0.5}$) 
\textbf{+Ours} & 98.2 & \textbf{95.5}\; \textbf{\textcolor{greencolor}{+5.0}} & \textbf{80.5}\; \textbf{\textcolor{greencolor}{+59.6}} & \textbf{58.8}\; \textbf{\textcolor{greencolor}{+52.7}} & \textbf{78.3}\; \textbf{\textcolor{greencolor}{+39.1}}  & \textbf{82.8} & 76.5\; 
\textcolor{greencolor}{+4.3}& \textbf{57.3}\; \textbf{\textcolor{greencolor}{+46.1}}& \textbf{35.8}\; \textbf{\textcolor{greencolor}{+32.9}}& \textbf{56.5}\; \textbf{\textcolor{greencolor}{+27.7}} \\
\bottomrule
\end{tabular}
\vspace{-5mm}
\end{center}
\end{table*}

%% file: table/ablation.tex
\begin{table}[t]
\caption{SUCCESS RATES (\%) FOR WRIST-VIEW ADAPTATION ABLATIONS ON LIBERO.
}
\setlength{\tabcolsep}{3pt} 
\label{ablation}
\begin{center}
% \begin{tabular}{c|c|ccc|c|c|ccc|c}
% \toprule
% \multirow{3}{*}{Baselines}
% & \multicolumn{5}{c|}{LIBERO}
% & \multicolumn{5}{c}{LIBERO-Plus} \\
% \cmidrule{2-6} \cmidrule{7-11}
% & \multirow{2}{*}{~Canonical~} & \multicolumn{4}{c|}{Perturbed} & \multirow{2}{*}{~Canonical~} & \multicolumn{4}{c}{Perturbed} \\
% \cmidrule{3-6} \cmidrule{8-11}
% &  & ~~~Small~~ & ~Medium~ & ~~Large~~~ & ~Average~ &  & ~~~Small~~ & ~Medium~ & ~~Large~~~ & ~Average~ \\    
% \midrule
% $\pi_{0.5}$ & 97.4 & 84.5 & 17.1 & 10.0 & 37.2 & 82.7 & 72.2 & 11.2 & 2.9 & 28.8 \\
% w/o MoE & 94.1 & 93.4 & 77.1 & 53.8 & 74.8 & 79.3 & 75.3 & 57.0 & 34.4 & 55.6 \\
% w/o Router & 95.5 & 91.8 & 77.5 & 54.3 & 74.5 & 81.2 & 74.5 & 56.3 & 34.5 & 55.1 \\
% w/o Anchor & - & - & - & - & - & 82.5 & 74.8 & 55.0 & 36.6 & 55.5 \\
% w/o Residual &  &  &  &  &   &  &  &  &  &  \\
% \midrule
% $\pi_{0.5}$ + Ours & 95.9 & 95.7 & 83.9 & 67.5 & 82.5 & 82.5 & 74.9 & 55.0 & 36.3 & 55.4 \\
% \bottomrule
% \end{tabular}
\begin{tabular}{c|c|ccc|c}
\toprule
\multirow{3}{*}{Baselines\vspace{-10pt}}
& \multicolumn{5}{c}{LIBERO} \\
\cmidrule{2-6}
& \multirow{2}{*}{~Canonical~\vspace{-6pt}} & \multicolumn{4}{c}{Perturbed}\\
\cmidrule{3-6}
&  & ~~~Small~~ & ~Medium~ & ~~Large~~~ & ~Average~ \\    
\midrule
% $\pi_{0.5}$ & 95.2 & 90.5 & 20.9 & 6.1 & 39.2 \\
w/o Experts & 94.1 & 93.4 & 77.1 & 53.8 & 74.8 \\
w/o Router & 95.5 & 91.8 & 77.5 & 54.3 & 74.5 \\
\midrule
$\pi_{0.5}$ + Ours & \textbf{98.2} & \textbf{95.5} & \textbf{80.5} & \textbf{58.8} & \textbf{78.3} \\
\bottomrule
\end{tabular}
\vspace{-5mm}
\end{center}
\end{table}

%% file: IEEEfull.bib
@string{ECCV = "Proceedings of the European Conference on Computer Vision (ECCV)"}

@string{AAAI = "Proceedings of the AAAI Conference on Artificial Intelligence (AAAI)"}

@string{ICLR = "Proceedings of the International Conference on Learning Representations (ICLR)"}

@string{NIPS = "Proceedings of the Neural Information Processing Systems (NeurIPS)"}

@string{ICRA = "Proceedings of the IEEE International Conference on Robotics and Automation (ICRA)"}

@string{CORL = "Proceedings of the Conference on Robot Learning (CoRL)"}

@string{RAL = "IEEE Robotics and Automation Letters (RA-L)"}

@string{RSS = "Robotics: Science and Systems (RSS)"}

@inproceedings{kim2024openvla,
    title={OpenVLA: An Open-Source Vision-Language-Action Model},
    author={{Moo Jin} Kim and others},
    booktitle=CORL,
    year={2024},
}

@InProceedings{pmlr-v305-black25a,
  title = 	 {$\pi_{0.5}$: a Vision-Language-Action Model with Open-World Generalization},
  author =       {Black, Kevin and others},
  booktitle = CORL,
  year = {2025},
}

@InProceedings{kim2025fine,
  title={Fine-Tuning Vision-Language-Action Models: Optimizing Speed and Success},
  author={Kim, Moo Jin and Finn, Chelsea and Liang, Percy},
  booktitle={arXiv preprint arXiv:2502.19645},
  year={2025}
}

@inproceedings{bjorck2025gr00t,
  title={Gr00t n1: An open foundation model for generalist humanoid robots},
  author={Bjorck, Johan and others},
  booktitle={arXiv preprint arXiv:2503.14734},
  year={2025}
}

@InProceedings{pmlr-v164-valassakis22a,
  title ={Learning Eye-in-Hand Camera Calibration from a Single Image},
  author ={Valassakis, Eugene and Dreczkowski, Kamil and Johns, Edward},
  booktitle = CORL,
  year = 	 {2022},
}

@InProceedings{pmlr-v305-wen25b,
  title = 	 {DexVLA: Vision-Language Model with Plug-In Diffusion Expert for General Robot Control},
  author =       {Wen, Junjie and Zhu, Yichen and Li, Jinming and Tang, Zhibin and Shen, Chaomin and Feng, Feifei},
  booktitle = CORL,
  year = 	 {2025},
}

@InProceedings{team2025gemini,
  title={Gemini robotics: Bringing ai into the physical world},
  author={Team, Gemini Robotics and others},
  booktitle={arXiv preprint arXiv:2503.20020},
  year={2025}
}

@InProceedings{li2024cogact,
  title={Cogact: A foundational vision-language-action model for synergizing cognition and action in robotic manipulation},
  author={Li, Qixiu and others},
  booktitle={arXiv preprint arXiv:2411.19650},
  year={2024}
}

@InProceedings{shukor2025smolvla,
  title={Smolvla: A vision-language-action model for affordable and efficient robotics},
  author={Shukor, Mustafa and others},
  booktitle={arXiv preprint arXiv:2506.01844},
  year={2025}
}

@inproceedings{rt12022arxiv,
    title={RT-1: Robotics Transformer for Real-World Control at Scale},
    author={Anthony	Brohan and others},
    booktitle={arXiv preprint arXiv:2212.06817},
    year={2022}
}

@inproceedings{rt22023arxiv,
    title={RT-2: Vision-Language-Action Models Transfer Web Knowledge to Robotic Control},
    author={Brohan, Anthony and others},
    booktitle={arXiv preprint arXiv:2307.15818},
    year={2023}
}

@inproceedings{octo_2023,
    title={Octo: An Open-Source Generalist Robot Policy},
    author = {Dibya Ghosh and others},
    booktitle = RSS,
    year = {2024},
}

@inproceedings{black2024pi,
  title={pi0: A Vision-Language-Action Flow Model for General Robot Control},
  author={Black, Kevin and others},
  booktitle={arXiv preprint arXiv:2410.24164},
  year={2024}
}

@inproceedings{jangir2022look,
  author={Jangir, Rishabh and Hansen, Nicklas and Ghosal, Sambaran and Jain, Mohit and Wang, Xiaolong},
  title={Look Closer: Bridging Egocentric and Third-Person Views with Transformers for Robotic Manipulation}, 
  booktitle=RAL, 
  year={2022}
}

@inproceedings{chen2024roviaug,
      title={RoVi-Aug: Robot and Viewpoint Augmentation for Cross-Embodiment Robot Learning},
      author={Lawrence Yunliang Chen and others},
      booktitle=CORL,
      year = {2024},
}

@inproceedings{tian2024vista,
    title   = {View-Invariant Policy Learning via Zero-Shot Novel View Synthesis},
    author  = {Stephen Tian and others},
    booktitle = CORL,
    year    = {2024},
}

@inproceedings{jiang2026knowyourcamera,
  title     = {Do You Know Where Your Camera Is? {V}iew-Invariant Policy Learning with Camera Conditioning},
  author    = {Tianchong Jiang and others},
  booktitle = ICRA,
  year = {2026},
}

@inproceedings{deng2025stereovla,
  title={Stereovla: Enhancing vision-language-action models with stereo vision},
  author={Deng, Shengliang and others},
  booktitle={arXiv preprint arXiv:2512.21970},
  year={2025}
}

@inproceedings{zhang2026grounding,
  title={Grounding actions in camera space: Observation-centric vision-language-action policy},
  author={Zhang, Tianyi and Duan, Haonan and Hao, Haoran and Qiao, Yu and Dai, Jifeng and Hou, Zhi},
  booktitle=AAAI,
  year={2026}
}

@inproceedings{li2026fixed,
  title={From Fixed to Free Cameras: Calibration-Free View-Robust Vision-Language-Action Model},
  author={Li, Wenhao and others},
  booktitle={arXiv preprint arXiv:2607.05396},
  year={2026}
}

@inproceedings{heo2026anycamvla,
  title={AnyCamVLA: Zero-Shot Camera Adaptation for Viewpoint Robust Vision-Language-Action Models},
  author={Heo, Hyeongjun and others},
  booktitle={arXiv preprint arXiv:2603.05868},
  year={2026}
}

@inproceedings{qu2025spatialvla,
  title={Spatialvla: Exploring spatial representations for visual-language-action model},
  author={Qu, Delin and others},
  booktitle={arXiv preprint arXiv:2501.15830},
  year={2025}
}

@article{li2026pointvla,
  title={Pointvla: Injecting the 3d world into vision-language-action models},
  author={Li, Chengmeng and Wen, Junjie and Peng, Yaxin and Peng, Yan and Zhu, Yichen},
  journal=RAL,
  year={2026},
}

@inproceedings{sun2025geovla,
  title={Geovla: Empowering 3d representations in vision-language-action models},
  author={Sun, Lin and Xie, Bin and Liu, Yingfei and Shi, Hao and Wang, Tiancai and Cao, Jiale},
  booktitle={arXiv preprint arXiv:2508.09071},
  year={2025}
}

@inproceedings{fan2026any3d,
  title={Any3d-vla: Enhancing vla robustness via diverse point clouds},
  author={Fan, Xianzhe and others},
  booktitle={arXiv preprint arXiv:2602.00807},
  year={2026}
}

@inproceedings{yang2026understanding,
  title={Understanding the Impact of Geometric Foundation Models on Vision-Language-Action Models},
  author={Yang, Yurou and Lin, Muyuan and Martin-Martin, Roberto and Labrie, Martin and Gayaka, Shreekant and Kuo, Cheng-Hao and Carlone, Luca},
  booktitle={ECCV},
  year={2026},
}

@inproceedings{li2026spatial,
  title={Spatial forcing: Implicit spatial representation alignment for vision-language-action model},
  author={Li, Fuhao and others},
  booktitle=ICLR,
  year={2026}
}

@inproceedings{abouzeid2025geoaware,
  title={Geoaware-vla: Implicit geometry aware vision-language-action model},
  author={Abouzeid, Ali and Mansour, Malak and Sun, Qinbo and Sun, Zezhou and Song, Dezhen},
  booktitle={arXiv preprint arXiv:2509.14117},
  year={2025}
}

@inproceedings{peng2026g,
  title={G3 VLA: Geometric inductive bias for Vision-Language-Action Models},
  author={Peng, Yue and others},
  booktitle={arXiv preprint arXiv:2606.24472},
  year={2026}
}

@inproceedings{han2026geometric,
  title={Geometric Action Model for Robot Policy Learning},
  author={Han, Jisang and others},
  booktitle={arXiv preprint arXiv:2606.17046},
  year={2026}
}

@inproceedings{jacobs1991adaptive,
  title={Adaptive mixtures of local experts},
  author={Jacobs, Robert A and Jordan, Michael I and Nowlan, Steven J and Hinton, Geoffrey E},
  booktitle={Neural computation},
  year={1991},
}

@inproceedings{hu2022lora,
    title={Lo{RA}: Low-Rank Adaptation of Large Language Models},
    author={Edward J Hu and others},
    booktitle=ICLR,
    year={2022}
}

@inproceedings{ebrahimpour2008view,
  title={View-independent face recognition with mixture of experts},
  author={Ebrahimpour, Reza and Kabir, Ehsanollah and Esteky, Hossein and Yousefi, Mohammad Reza},
  booktitle={Neurocomputing},
  year={2008},
}

@inproceedings{jin2024lvsm,
  title={Lvsm: A large view synthesis model with minimal 3d inductive bias},
  author={Jin, Haian and others},
  booktitle=ICLR,
  year={2025}
}

@inproceedings{liu2023libero,
  title={Libero: Benchmarking knowledge transfer for lifelong robot learning},
  author={Liu, Bo and others},
  booktitle=NIPS,
  year={2023}
}

@inproceedings{fei25libero,
    title={LIBERO-Plus: In-depth Robustness Analysis of Vision-Language-Action Models},
    author={Senyu Fei and others},
    booktitle = {arXiv preprint arXiv:2510.13626},
    year={2025},
}

@inproceedings{dosovitskiy2020image,
  title={An image is worth 16x16 words: Transformers for image recognition at scale},
  author={Dosovitskiy, Alexey and others},
  booktitle=ICLR,
  year={2021}
}

@inproceedings{kingma2014adam,
  title={Adam: A method for stochastic optimization},
  author={Kingma, Diederik P and Ba, Jimmy},
  booktitle=ICLR,
  year={2015}
}

@inproceedings{welle2024quest2ros,
title={Quest2ROS: An App to Facilitate Teleoperating Robots},
author={Welle, Michael C and others},
booktitle={7th International Workshop on Virtual, Augmented, and Mixed-Reality for Human-Robot Interactions},
year={2024} }
